\documentclass{MITcsail}
\usepackage{float}
\usepackage{placeins}
\usepackage{algorithm}
\usepackage{algpseudocode}
\algrenewcommand{\algorithmicrequire}{\textbf{Input:}}
\algrenewcommand{\algorithmicensure}{\textbf{Output:}}
\usepackage[most]{tcolorbox}
\definecolor{PromptField}{RGB}{242,90,40}
\newcommand{\promptfield}[1]{\textcolor{PromptField}{\textbf{#1:}}}
\newtcolorbox{promptbox}[1]{
  enhanced, breakable,
  colback=black!3, colframe=black!20,
  boxrule=0.6pt, arc=2mm,
  left=3mm, right=3mm, top=3mm, bottom=3mm,
  title={#1}, fonttitle=\small,
  colbacktitle=black!50, coltitle=white,
  attach boxed title to top left={xshift=4mm,yshift=-2mm},
  boxed title style={boxrule=0pt,arc=0.5mm},
  fontupper=\small
}
\newcommand{\pagehead}{PCBRouteAgent}

\title{PCBRouteAgent: How Close Are LLM Agents to Becoming PCB Routing Engineers?}

\author[1]{Yue Zhang$^*$}
\author[1]{Jilei Sun$^*$}
\author[1]{Liqiang Jing$^{*,\dagger}$}
\author[2]{Yijie Wei$^\ddagger$}
\author[3]{Ruohan Yang}

\affil[1]{The University of Texas at Dallas}
\affil[2]{Texas Instruments Inc.}
\affil[3]{Iowa State University}

\begin{document}

\maketitle
\noindent{\small $^*$\ Equal contribution.\quad $^\dagger$\ Corresponding author:
Liqiang Jing (\href{mailto:jingliqiang6@gmail.com}{jingliqiang6@gmail.com}).\\
$^\ddagger$\ This work was conducted outside the author's role at Texas Instruments Inc.\par}
\thispagestyle{firstpagestyle}

\begin{abstract}
Automating printed circuit board (PCB) routing requires coordinating board-level connectivity decisions with geometric execution under strict design rules. Although recent agents can interact with native routing tools, reliable routing on complex boards remains challenging. We present \textbf{PCBRouteAgent}, a multimodal PCB routing agent inspired by how human engineers coordinate global strategy with local execution. A Global Routing Planner combines board images and structured electrical and geometric information to specify connection topology, routing order, corridors, and layer transitions. A Detailed Routing Executor realizes these guides through KiCad's native routing engine and performs bounded local recovery. Unresolved failures are returned as structured feedback, enabling targeted guide revisions while preserving accepted routes. We evaluate PCBRouteAgent on 247 synthetic and real-world boards from four PCBWorld-Bench test splits. It achieves clean-pass rates of 100\%, 94.9\%, 100\%, and 40\% on D2, D3-A, D3-B, and D3-C, respectively. On the three real-world subsets, PCBRouteAgent outperforms the best evaluated baseline in clean-pass rate by 13.9, 36, and 10 percentage points, respectively.
Case studies reveal limitations in routing-order planning and signal-specific constraints, highlighting a substantial remaining gap between connection completion and electrically validated PCB designs.
The project page and code are available at \url{https://chip-ai-agent.github.io/research/pcbrouteagent} and \url{https://github.com/chip-ai-agent/PCBRouteAgent}, respectively.
\end{abstract}

\section{Introduction}
\begin{figure}
    \centering
    \includegraphics[width=0.8\linewidth]{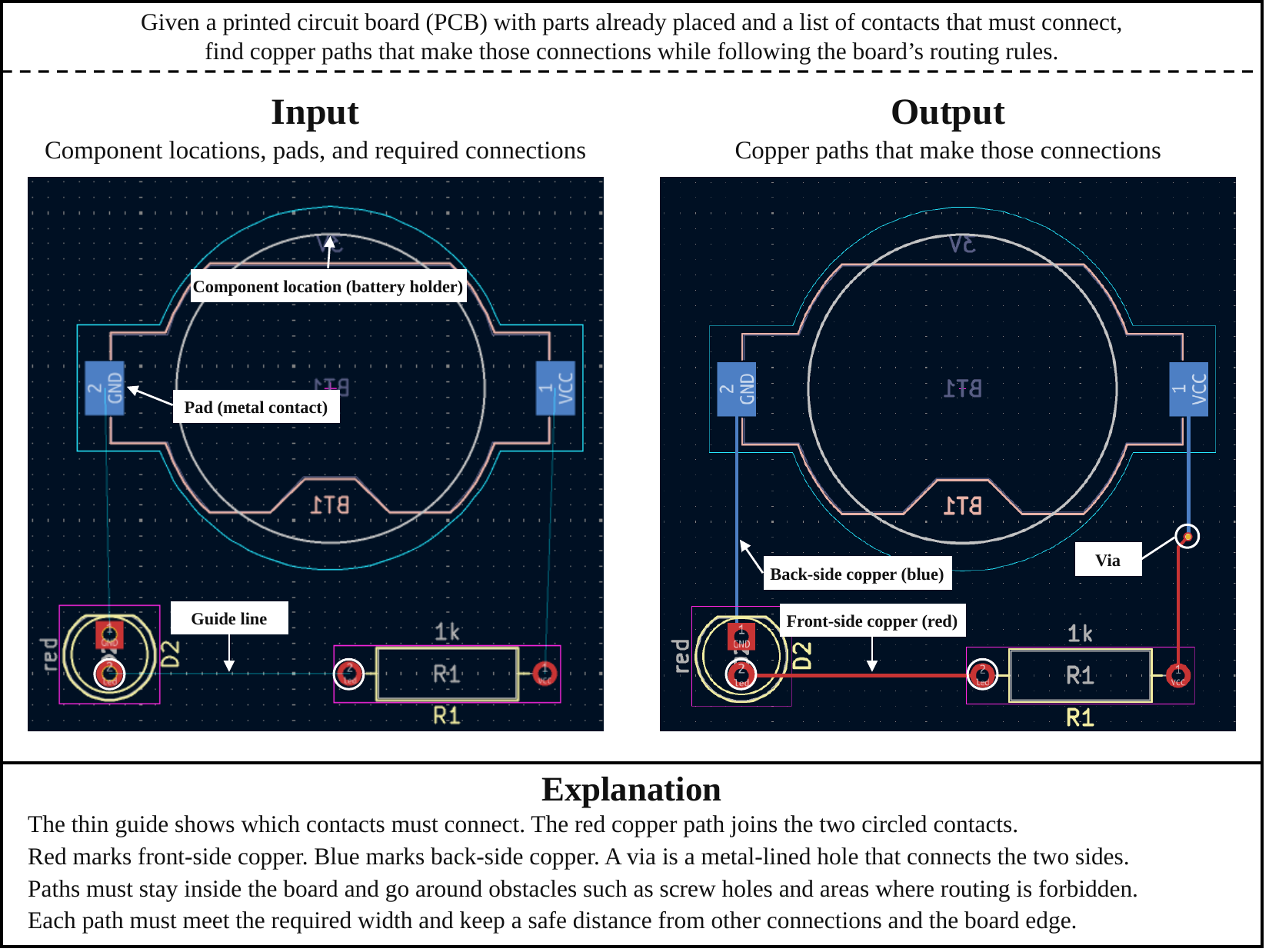}
    \caption{PCB routing connects required pads with copper traces and vias while satisfying design rules.}
    \label{fig:pcb-routing-task}
\end{figure}

Printed circuit boards (PCBs) provide mechanical support and electrical interconnections for electronic components. Before a circuit design can be fabricated as a PCB, its schematic connectivity must be translated into a physical board layout. This process includes component placement, which determines the positions and orientations of component footprints, followed by routing, which realizes the required electrical connections. As illustrated in Figure~\ref{fig:pcb-routing-task}, routing connects the pads belonging to each electrical net\footnote{A net is an electrical connectivity group in a PCB design, specifying the component terminals that must be electrically connected according to the circuit schematic.} through copper traces and, when necessary, vias that enable transitions between copper layers. A net may contain two or more pads, all of which must form a connected network while remaining electrically isolated from other nets. The resulting routing must satisfy design rules governing clearance, trace width, vias, and board boundaries, as well as signal-specific constraints such as differential-pair requirements~\cite{lin2021unified,song2026pcbworld}. Beyond establishing valid connections, routing also seeks to limit wirelength and unnecessary layer transitions while preserving space for other nets. These requirements make PCB routing a coupled physical design problem in which each routing decision affects the connections that remain to be completed.

Routing complex PCBs remains labor-intensive, motivating automatic routers that reduce the need for manual trace construction. Classical rule-based methods use heuristic search~\cite{lee1961path}, net ordering, and congestion-aware costs~\cite{lin2021unified} to construct routes under geometric constraints. Although these methods automate much of the process, complex boards can still require careful tuning and manual intervention~\cite{song2026pcbworld}. To reduce this reliance on handcrafted decisions, learning-based approaches have been explored for routing tasks, including global routing~\cite{liao2020global}, net ordering~\cite{liao2026worldmodel}, and PCB fanout~\cite{li2023fanout}. Their effectiveness, however, depends on how routing decisions are represented and connected to physical execution. As discussed in PCBWorld~\cite{song2026pcbworld}, grid-based formulations face rapidly growing decision spaces as routing resolution increases, while approaches that guide an existing autorouter leave much of the physical routing process to that router. These limitations motivate studying agents that interact directly with a design engine and adapt their decisions using execution feedback. PCBWorld~\cite{song2026pcbworld} provides such an environment by exposing KiCad's native routing operations and design-rule feedback to reinforcement-learning policies and tool-using LLM agents. This interaction allows agents to observe the consequences of routing decisions and revise subsequent actions as the board evolves.

Despite this progress, reliable agent-based PCB routing remains challenging. The LLM agents evaluated in PCBWorld~\cite{song2026pcbworld} still lag behind rule-based routers in completing connections while satisfying design rules, particularly on complex real-world boards. Providing access to native routing operations and execution feedback does not, by itself, ensure effective board-level coordination. An agent must decide which connections to prioritize, allocate limited routing space across competing nets, and determine when a layer transition is necessary. These decisions are coupled: a locally feasible route may block subsequent connections, while a connection that repeatedly fails may require a change in its topology or layer assignment rather than another local attempt. Moreover, as routing proceeds, the agent must distinguish difficulties that can be resolved locally from those requiring a broader strategy revision. These challenges highlight the need for an explicit mechanism that coordinates board-level planning with geometric execution and uses routing failures to revise the plan.

To address these limitations, we propose \textbf{PCBRouteAgent}, a multimodal PCB routing agent inspired by how human routing engineers coordinate global strategy with local execution. Engineers consider connection priorities, routing regions, and layer usage while constructing individual traces, and revisit these decisions when local obstacles prevent further progress. Following this principle, PCBRouteAgent integrates a \textbf{Global Routing Planner} and a \textbf{Detailed Routing Executor} through a feedback-driven planning--execution loop. The planner combines board images with structured electrical and geometric information to organize connectivity demands into routing guides specifying connection topology, routing order, corridors, and layer transitions. The executor translates these guides into physical traces and vias through KiCad's native routing engine, using bounded local recovery to handle geometric difficulties. When recovery is insufficient, it returns structured execution evidence to the planner, which generates targeted guide revisions while preserving accepted routes. This design enables board-level guidance to adapt to the physical constraints revealed during execution.

We evaluate PCBRouteAgent on PCBWorld-Bench~\cite{song2026pcbworld}, comprising 128 synthetic boards and 119 real-world boards across four test splits. PCBRouteAgent achieves a 100\% clean-pass rate on D2. On the real-world D3-A, D3-B, and D3-C subsets, PCBRouteAgent achieves clean-pass rates of 94.9\%, 100\%, and 40\%, respectively, exceeding the best evaluated baseline on each split by 13.9, 36, and 10 percentage points. These baselines achieve 81\%, 64\%, and 30\% clean pass: PCBWorld PPO w/o finish~\cite{song2026pcbworld} on D3-A, and Freerouting~\cite{freerouting2026} on D3-B and D3-C. It also outperforms the published PCBWorld GPT-5.4 agent~\cite{song2026pcbworld,openai2026gpt54}, whose clean-pass rates are 42\% on D3-A and 0\% on D3-B. On D3-C, PCBRouteAgent reaches 94.9\% mean routability and reduces the average design-rule violation count from Freerouting's 32.0 to 6.3 per board. Despite these gains, our case studies reveal important limitations (Appendix~\ref{app:case-studies}). Earlier routing decisions can block subsequent connections, while fully connected outputs can overlook differential-pair relationships. A substantial gap therefore remains between strong benchmark performance and the broader requirements of PCB engineering.

\section{Related Work}

\subsection{PCB Routing}
PCB routing connects the pads specified by a netlist while respecting the
geometry of components, existing copper, board boundaries, and electrical
design rules. Classical approaches search for feasible paths and then refine
them according to costs such as wirelength, congestion, and the number of vias. In IC routing, TritonRoute~\cite{kahng2021tritonroute} illustrates the global--detailed decomposition: global routing determines coarse topology and layer guidance, while detailed routing realizes traces and vias under geometric design rules.
Lee's maze-routing algorithm~\cite{lee1961path} established a systematic way to find connections
in a discretized search space; subsequent PCB routers have
incorporated constraints that simple shortest-path search cannot capture. For
example, Lin et al.~\cite{lin2021unified} address complicated board constraints and differential
pairs within a unified routing algorithm. These methods
demonstrate the value of explicit geometric search and rule-aware optimization,
but they also make clear that routing quality depends on decisions made across
multiple scales: the order in which nets are attempted, the space reserved for
future connections, and the local shape of each trace. Learning-based work has
targeted different parts of this decision process. Liao et al.~\cite{liao2020global} investigate
deep reinforcement learning for global routing, while
FanoutNet~\cite{li2023fanout} learns decisions for the specialized PCB fanout
problem. GPCB~\cite{chen2025gpcb} uses a generative pretrained Transformer to predict
routing-related network-flow patterns and then constructs physical routes from
those predictions. More recently, world-model reinforcement
learning~\cite{liao2026worldmodel} has been integrated with Freerouting so that learned guidance can be
combined with an existing autorouter. These systems
differ in what they predict and in how much of the routing procedure is handled
by a separate search or routing engine. PCBWorld~\cite{song2026pcbworld} provides another setting: it
exposes KiCad's native routing operations to reinforcement-learning policies
and tool-using LLM agents, and evaluates the resulting board with engine-based
design-rule checks. This interactive setting makes it
possible to observe how a routing decision changes the board before choosing
the next action. Our work uses that setting to study the agent's ability to
coordinate routing decisions over an evolving board, rather than treating each
connection as an isolated path-finding problem.

\subsection{LLM Agents for EDA}
LLM agents have begun to use EDA tools as part of longer design workflows,
where a model plans tool calls and uses their feedback to guide later steps. ReAct~\cite{yao2023react} interleaves reasoning with environment actions, while Reflexion~\cite{shinn2023reflexion} uses feedback and reflective memory to improve subsequent attempts without updating model weights.
Such workflows require the agent to track design state, choose suitable tools,
and recover when a tool reports an error. ChatEDA~\cite{wu2024chateda} connects language-model planning
to EDA tool invocation in an RTL-to-GDSII flow, and EDAid~\cite{wu2025edaid}
investigates collaboration among multiple agents for electronic design
automation. VerilogCoder~\cite{ho2025verilogcoder} focuses on HDL development, using
planning and feedback from code-oriented tools, including waveform tracing,
to revise generated Verilog. Although these systems
address different stages of hardware design, they share a useful pattern:
the agent proposes an action, receives evidence from a domain tool, and uses
that evidence to decide what to do next. PCB layout presents a different
physical interface to this pattern. PCBAgent~\cite{chen2025pcbagent} combines an RL-based placement
agent with an LLM agent for interactive component-placement optimization; placement choices affect routability, but placement
alone does not test whether an agent can complete the required copper
connections. Recent PCB benchmarks have made such distinctions more visible.
PCB-Bench~\cite{li2026pcbbench} evaluates LLMs on PCB placement and routing tasks, including the
interpretation of board information. OmniRouting~\cite{lu2026omnirouting} assesses
routing reasoning under geometric, design-rule, and electrical constraints,
and includes a tool-augmented task. These evaluations
help characterize what a model understands or proposes, while an interactive
routing environment additionally reveals whether a sequence of actions yields
a valid board as the geometry changes. PCBWorld~\cite{song2026pcbworld} supplies this engine-grounded
interface and evaluates completed boards with native design-rule feedback. PCBRouteAgent therefore studies an LLM agent in the
closed loop between planning, tool execution, board-state observation, and
validation. The emphasis is on the agent's ability to maintain global routing
objectives while adapting its next decision to the board produced by earlier
actions.

% Methods revised against the user inline feedback.
% Requires \usepackage{graphicx}; section numbering follows the parent manuscript.
% Figure artwork is unchanged; overview uses the recovered first original.

\section{Methods}
\label{sec:method}

From the perspective of a PCB routing engineer, routing requires decisions at two spatial scales: coordinating connection priorities, routing regions, and layer usage across the board, and realizing individual traces around pads and obstacles. A locally feasible connection may consume space needed by another net, while a promising board-level strategy may prove geometrically infeasible during execution. This motivates separating global planning from detailed routing and revising the plan when local difficulties arise. A similar separation is used in IC routing flows such as OpenROAD~\cite{openroadRouting}, where global routing produces guides for detailed routing.
% We adopt this principle for PCB routing: the Global Routing Planner specifies connection topology, routing order, corridors, and layer transitions, while the Detailed Routing Executor constructs physical copper and returns feedback for plan revision.

\begin{figure}[!t]
  \centering
  \includegraphics[width=\linewidth]{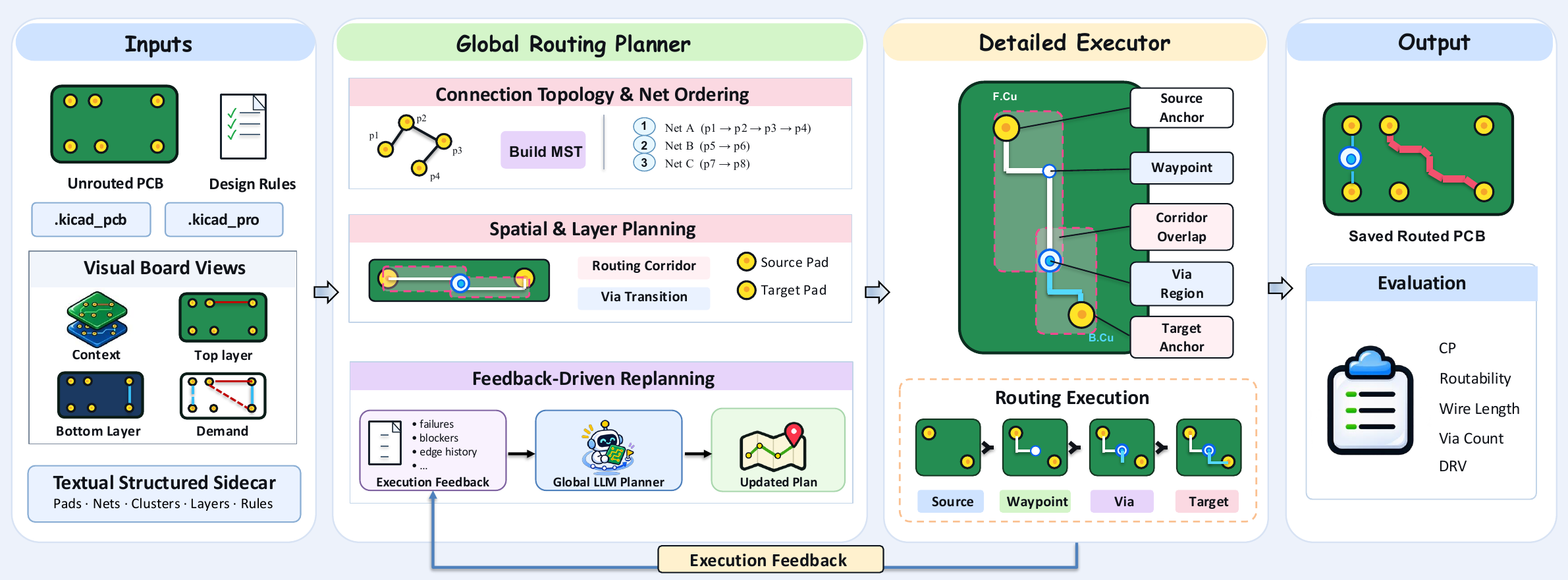}
  \caption{Overview of PCBRouteAgent.}
  \label{fig:routeagent}
\end{figure}

We adopt this principle for PCB routing. As illustrated in Figure~\ref{fig:routeagent}, PCBRouteAgent is a multimodal agentic framework for PCB routing that integrates two tightly coupled components: a Global Routing Planner and a Detailed Routing Executor. The Global Routing Planner takes on the board-level planning responsibilities of a human PCB routing engineer, developing and maintaining a high-level routing strategy that covers key decisions such as connectivity planning, routing priorities, and spatial and layer planning. Guided by this strategy, the Detailed Routing Executor establishes physical connections through KiCad's native push-and-shove (PNS) router~\cite{kicad2026pcbnew,song2026pcbworld} while handling local obstacles and geometric constraints, generating structured execution feedback when routing attempts fail. The Global Routing Planner then leverages this feedback and multimodal board context to revise its strategy and guide subsequent execution, enabling the agent to iteratively adapt its routing decisions. The overall procedure is summarized in Algorithm~\ref{alg:pcbrouteagent} in Appendix~\ref{app:routing-algorithm}.

\subsection{Task Formulation}
\label{sec:task-formulation}
Given an initial PCB $B_0$, a target net set $\mathcal{N}$, and design rules $\mathcal{R}$, the routing task is to produce a routed board $B_r = \mathcal{F}(B_0, \mathcal{N}, \mathcal{R})$, where $\mathcal{F}$ denotes the routing agent that maps the input board, target nets, and design rules to the resulting board. Here, $B_0$ specifies fixed component placements, pad geometries, copper layers, obstacles, and existing routing; $\mathcal{N}$ specifies the required electrical connections; and $\mathcal{R}$ defines track widths, clearances, and via constraints. The agent accesses these inputs through board images and structured board information. The output $B_r$ contains physical traces and vias saved in a KiCad PCB file. The objective is to connect all pads belonging to each target net while satisfying the design rules and keeping wirelength and via count low.

\subsection{Global Routing Planner}
\label{sec:global-agent}
Routing multiple nets on a shared PCB requires coordinating limited routing space while maintaining electrical connectivity. To address this challenge, the Global Routing Planner first organizes disconnected components within each net into a connection topology and then assigns spatial routing guides to support their physical realization. The resulting routing guide provides the Detailed Routing Executor with a structured plan for establishing these connections.

% 输入
% 1. Board context元件信息 placement 上下层都有的 以截图形势给出
% 四张图片 2.顶层元件placement 图片
% 3. 底层placement
% 4. Demand 告诉你net list 总的

% Structured text
% 跟PCBWorld一样 文字形式 net net里面的焊点坐标 线的宽度 这根线是什么线 用来干嘛的？ 板子的信息 routing geometry

% 输出 corridor + 穿孔 via json文件 结构化信息 xxx

\subsubsection{PCB BoardRepresentation}
\label{sec:board-representation}
Effective global routing requires understanding both the outstanding connectivity demands and the physical constraints under which these connections must be established. Given the initial PCB $B_0$, target net set $\mathcal{N}$, and design rules $\mathcal{R}$, we construct a multimodal board representation comprising \textbf{board images} $\mathcal{I}$ and \textbf{structured text} $S$. Through PCBWorld's KiCad-based engine~\cite{song2026pcbworld,kicad2026pcbnew}, we extract pad geometries, copper layers, obstacles, and existing routing from $B_0$, target-net connectivity from $\mathcal{N}$, and routing constraints from $\mathcal{R}$. This information is represented in two complementary forms. The \textbf{visual modality} $\mathcal{I}=\{I_{\mathrm{layout}},I_{\mathrm{front}},I_{\mathrm{back}},I_{\mathrm{demand}}\}$ depicts the overall board layout, front- and back-copper states, and outstanding connectivity demands. The \textbf{textual modality} $S$ serializes pad coordinates and geometries, layer information, net assignments, connectivity clusters, and design rules. For each target net $n\in\mathcal{N}$, electrically connected pads form clusters $\mathcal{C}_n=\{C_{n,1},\ldots,C_{n,K_n}\}$, whose remaining connections define the routing demands. Together, the images provide spatial context, while the structured text supplies precise identifiers and numerical constraints for global planning.

\subsubsection{Connection Topology and Routing Order}
Although the board representation identifies the remaining connectivity demands, it does not specify how disconnected clusters should be connected. For a net containing multiple clusters, numerous connection topologies can satisfy the same electrical requirements, while independently connecting every pair would introduce unnecessary routing. To avoid exhaustively searching this combinatorial space, we formulate connectivity planning as a minimum spanning tree (MST)~\cite{prim1957shortest} problem over the disconnected clusters, yielding a compact initial topology. For each target net $n$, we construct a weighted graph $G_n=(\mathcal{C}_n,\mathcal{E}_n,w_n)$, where vertices represent connectivity clusters and edges represent candidate inter-cluster connections. Each candidate edge $e\in\mathcal{E}_n$ is assigned a weight $w_n(e)$ based on the minimum Euclidean distance between pads in the corresponding clusters, with an additional penalty for incompatible pad layers. The resulting MST $T_n=(\mathcal{C}_n,\mathcal{E}_n^*)$ specifies $K_n-1$ connection tasks spanning the $K_n$ clusters. It minimizes the estimated total cost under this heuristic, while leaving the physical realization of these connections to subsequent routing stages.

Beyond determining the connection topology within each net, the planner must also coordinate routing across multiple nets that share limited board space. We therefore initialize the routing order using the total estimated connection length of each net's spanning tree, prioritizing shorter nets first. This provides a lightweight starting strategy for organizing routing tasks, without requiring an exhaustive search over possible execution orders. Both the connection topology and routing priorities remain adjustable as subsequent execution reveals routing constraints not captured by the initial estimates.

\subsubsection{Spatial and Layer Planning}
While the connection topology specifies which clusters should be connected, it does not determine how these connections should traverse the physical board. Directly prescribing exact trace geometries at the global planning stage would require resolving fine-grained obstacles and design-rule constraints before routing execution. Instead, we separate high-level spatial guidance from low-level geometric realization, allowing the Global Routing Planner to guide routing decisions while preserving flexibility for local obstacle avoidance. Specifically, for each connection edge $e$, we use the endpoint pads selected during topology construction to initialize ordered corridor guidance $\mathcal{Q}_e=(\mathcal{Q}_{e,1},\ldots,\mathcal{Q}_{e,m_e})$, where $m_e$ is the number of corridor stages for connection $e$. When the endpoints share a copper layer, we construct a single corridor $\mathcal{Q}_{e,1}$ from their axis-aligned bounding box, expanded by a margin $\delta$ in each direction. The resulting rectangle provides spatial guidance while leaving the exact trace geometry to the Detailed Routing Executor.

Beyond spatial guidance, routing may also require transitions between copper layers when the endpoints cannot be connected on a common layer. We split the connection at its midpoint and construct two layer-specific corridors, $\mathcal{Q}_{e,1}$ from the source pad to the midpoint and $\mathcal{Q}_{e,2}$ from the midpoint to the target pad, using the same bounding-box expansion. A via-transition region $\mathcal{V}_e$ is initialized by expanding the midpoint by $\delta$ in each direction. Rather than fixing an exact via location, the planner specifies a candidate region in which the Detailed Routing Executor can search for a feasible transition using KiCad's native routing engine. The resulting connection topology, routing order, and spatial and layer guidance are consolidated into an initial routing guide, which is validated and translated into routing tasks for the Detailed Routing Executor.

\subsubsection{Feedback-Driven Replanning}
\label{sec:feedback-replanning}
Although heuristic initialization provides an efficient starting point, its simplified cost estimates and spatial guidance cannot fully anticipate the geometric constraints encountered during physical routing. Connections that appear feasible at the planning stage may fail due to obstacles, restricted routing space, or layer constraints. To address this gap, the Global Routing Planner revises its routing strategy using evidence obtained from detailed execution.

When a connection remains unresolved after the Detailed Routing Executor's local recovery attempts (Section~\ref{sec:local-recovery}), the executor produces structured feedback $\mathcal{F}_e$ for each unresolved connection $e$, describing the failed attempts, last reached positions, potential obstacles, and deviations from the original guide. These feedback packets are aggregated and provided to the multimodal LLM used by the Global Routing Planner, together with the initial board images $\mathcal{I}$, structured board information $S$, the current routing guide, and updated connectivity summaries. The initial images remain fixed across replanning rounds; execution feedback and connectivity summaries provide observations of the evolving board. By jointly considering spatial context and execution evidence, the planner can revise high-level decisions that local routing alone cannot resolve.

At replanning round $t$, let $\mathcal{G}_t$ denote the current routing guide, $\mathcal{D}_t$ the batch of feedback packets $\mathcal{F}_e$ for unresolved connections, and $\mathcal{K}_t$ the updated connectivity summaries. The multimodal LLM $\mathcal{M}_{\theta}$ generates a guide patch, a structured set of local updates to the current routing guide, as
\begin{equation}
    \Delta\mathcal{G}_t = \mathcal{M}_{\theta}\bigl(
    \mathcal{I}, S, \mathcal{G}_t, \mathcal{D}_t, \mathcal{K}_t;
    P_{\mathrm{sys}}, P_{\mathrm{replan}}\bigr),
    \label{eq:llm-replanning}
\end{equation}
where $\theta$ denotes the model parameters, $\Delta\mathcal{G}_t$ is the structured output patch, and $P_{\mathrm{sys}}$ and $P_{\mathrm{replan}}$ are the system instructions and replanning prompt template, respectively. Their templates are listed in Appendices~\ref{app:prompt-global-system} and~\ref{app:prompt-replanning}.

The planner outputs a structured guide patch specifying targeted changes to connection endpoints, routing corridors, via-transition regions, or routing order. The patch updates the affected portions of the current guide while preserving previously accepted routes outside that scope. Before execution resumes, the patch is checked for consistency with the current snapshot and guide revision, as well as valid references to clusters, corridors, and copper layers. If validation fails, the same planner may be asked to correct its response using the previous output and validation errors; the correction prompt template is listed in Appendix~\ref{app:prompt-repair}. The validated patch is merged into the guide, and affected connection tasks are recompiled for the Detailed Routing Executor. This planning--execution loop continues until the target connections are completed, no further tasks can be scheduled, or the execution or replanning budget is exhausted.

% 目标保证1.所有点连接 2. 连接所有的点之间线的长度和最短 so 最小生成树 出事最优的拓扑

%  确定拓扑结构 -》不一定 联通
%  LLM 根据net内部的拓扑结构 读所有的net 拓扑  给走廊 corridor （两点之间的布线空间）（prompt）为了资源的分配？ 拥挤 走廊分配很关键 搜索空间减少 好处 带有efficient的好处

%  有了走廊 从分配的走廊里吗搜索不到可行路径 所以要穿孔分配 （）和上面是one step

% 继续detailed agent：PNS push and shove 现在是最简单原则最短优先 不用绕很多
% local retry： 跳过当前net / 丢掉 算各种metrics evaluator 具体啥问题

% 根据反馈改穿孔和corridor MST改没改？

% \begin{figure}[H]
%   \centering
%   \includegraphics[width=\textwidth]{fig/routeagent_representation.pdf}
%   \caption{Image and structured-text inputs describe the same real board in a shared physical coordinate frame. The example guide connects clusters anchored at CB1:1 and U3:1 using ordered B.Cu and F.Cu corridors and a via-transition region. Colored regions show planning guidance, rather than committed copper. The guide also specifies endpoint clusters, routing profiles, and net order. The input fields and initial guide come from the same archived snapshot.}
%   \label{fig:representation}
% \end{figure}

\subsection{Detailed Routing Executor}
While the Global Routing Planner determines which connections to establish and where they should be routed, realizing these decisions requires resolving precise geometric constraints that cannot be fully captured by high-level guidance. The Detailed Routing Executor translates the routing guide into physical traces and vias through KiCad's native PNS router, attempts bounded local recovery when routing fails, and returns unresolved failures to the Global Routing Planner for guide revision.

\subsubsection{Guided Routing Execution}

The Detailed Routing Executor takes the routing guide produced by the Global Routing Planner and compiles its planned connection edges into executable tasks. For each connection edge $e$, a task specifies the source and target connectivity clusters, candidate pad anchors, ordered layer-specific corridors $\mathcal{Q}_e$, and optional via-transition regions $\mathcal{V}_e$. The executor uses this guidance together with the design rules $\mathcal{R}$ to select routing targets and invoke native routing operations. These tasks are scheduled according to the routing priorities established by the Global Routing Planner, with connections within each net following their predefined task order. For each task, the executor selects a source anchor from the designated cluster that lies within the initial routing corridor and supports its assigned copper layer, prioritizing candidates closer to the next routing target. A compatible target anchor is identified from the destination cluster based on the final routing stage.

Instead of prescribing a fixed trace path, the executor progressively advances through the planned routing stages by selecting intermediate waypoints. For consecutive corridors $\mathcal{Q}_{e,j}$ and $\mathcal{Q}_{e,j+1}$ on the same layer, it identifies their overlapping region and generates candidate waypoints at the region center and surrounding positions, accounting for trace width and clearance requirements. These candidates are attempted sequentially through PNS, which constructs the physical trace between the current routing position and the selected target while handling local geometric constraints. Upon a successful attempt, the executor updates its routing position and proceeds to the next stage.

When a connection requires a layer transition, the executor instead generates candidate via locations within the designated transition region $\mathcal{V}_e$, accounting for via dimensions and clearance requirements. It then attempts to establish the transition before continuing on the next copper layer. After traversing the planned routing stages, the executor attempts to reach the target cluster and verifies whether the source and target clusters have become electrically connected. This connectivity-based verification ensures that task completion reflects an established physical connection rather than merely a successful intermediate routing operation.

\subsubsection{Local Recovery and Execution Feedback}
\label{sec:local-recovery}
Although PNS can handle many local geometric constraints, routing attempts may still fail when obstacles block the selected path, restrict access to intermediate waypoints, or prevent a layer transition. Immediately triggering global replanning for every failure would introduce unnecessary overhead, particularly when alternative local paths remain available. We therefore employ a bounded recovery strategy that progressively explores alternative routing options before escalating unresolved failures to the Global Routing Planner.

When a routing attempt fails, the executor first explores local geometric alternatives through PNS, using walkaround to navigate obstacles and shove to adjust obstructing traces when necessary. If the current target remains unreachable, it attempts alternative waypoints within the designated routing region. For more challenging same-layer obstacles, a grid-based A* search~\cite{hart1968formal} identifies potential detours, which are subsequently executed through PNS. The executor can further backtrack to earlier routing stages or switch to alternative source anchors within the same connectivity cluster. These operations expand the local search space while preserving the connection topology and layer-transition regions established by the Global Routing Planner. To avoid excessive exploration, recovery attempts are bounded by retry and execution budgets.

When local recovery is exhausted or a planned layer transition cannot be completed, the executor generates a structured feedback packet $\mathcal{F}_e$ for the current connection $e$, describing the attempted routing operations, last reached position, potential blocking obstacles, affected routing regions, and deviations from the original guidance. Failed tasks are rolled back to a consistent checkpoint, while previously accepted connections outside the affected scope are preserved. Rather than reporting failure as a binary outcome, the executor provides actionable evidence that helps the Global Routing Planner identify which higher-level decisions may require revision. This separation allows local geometric difficulties to be addressed through efficient recovery operations, while reserving global replanning for failures that cannot be resolved within the existing routing strategy. The resulting feedback is passed to the Global Routing Planner for the feedback-driven replanning described in Section~\ref{sec:feedback-replanning}.

\paragraph{Termination and output.}
The execution controller terminates the routing loop when all target connections are established, no further tasks or permitted recovery attempts remain, or the execution budget is exhausted. Connections that exhaust their replanning allowance are recorded as unresolved while other tasks may continue. A checkpoint or rollback failure also triggers early termination. On termination, the system saves the last consistent board state and reports completed and unresolved connections. Routing success is determined by the official evaluator applied to the saved board.

\section{Experimental Setup}

\paragraph{Benchmarks.}
We follow the agent-comparison setting of PCBWorld-Bench
~\cite{song2026pcbworld}. Each instance is a board in KiCad's~\cite{kicad2026pcbnew} native
\texttt{.kicad\_pcb} format, with reference routes removed; component
placement, netlist, and design rules remain fixed. The task is to complete the
connections on that board, and the final board is evaluated by KiCad rather
than by the agent's own estimate of success. We use the same three test sets as
the PCBWorld main comparison: 128 synthetic, two-layer gridless boards from
D2-test; 99 small open-source boards from D3-A; and 10 medium open-source
boards from D3-B. We additionally include the 10 large open-source boards in
PCBWorld's~\cite{song2026pcbworld} fixed D3-C evaluation subset. PCBWorld
excludes one of the 100 D3-A boards because its engine interface cannot load
it. The D3-B and D3-C test subsets contain two-layer boards selected across
each split's pad-count range; D3-C contains boards with more than 100 pads.
D2 tests routing on boards drawn from the synthetic distribution, while all
three D3 splits test transfer to real designs with their original design-rule
settings. The Reference rows in Tables~\ref{tab:routeagent-d3a}--\ref{tab:routeagent-d3c} evaluate the original routed designs retained in PCBWorld's open-source board corpus~\cite{song2026pcbworld}. Each routing input is obtained by removing the target routing while preserving component placement, the netlist, and design rules; the Reference is the retained design before this removal, scored using the same official engine-based evaluator. D3-A and D3-B Reference values are taken from PCBWorld's published Table~22, while the D3-C Reference values are evaluated locally. These rows provide an existing-design reference for comparison with automatically produced routes. We use PCBWorld's
published board identifiers and split definitions to keep the comparison
board-for-board.

\paragraph{Protocol and metrics.}
We use PCBWorld's KiCad-based board evaluator~\cite{song2026pcbworld} for
the reported routing quality. Published D2, D3-A, and D3-B baseline values
in Tables~\ref{tab:routeagent-d2}--\ref{tab:routeagent-d3b} use PCBWorld's
single-rollout (@1) means from Table~22. PCBRouteAgent contributes one scored
board per instance; any retries of unscored model-response failures are
reported separately from this evaluation convention. Clean Pass (CP) is the primary
success measure: a
board passes only if all required connections are complete and KiCad reports
zero error-level design-rule violations. We also report Potential Gain (Pot.),
the change in the benchmark's board-potential score from the unrouted state
to the final board; Routability (Rout.), the fraction of required connections
completed; and design-rule violation count (DRV), wirelength (WL), and via
count. The main tables report routing quality. We additionally compare runtime, model calls, token usage, and API cost for our local PCBWorld and PCBRouteAgent runs under the same evaluation conditions with Claude Haiku 5.5. Both methods use the same timing scope; detailed efficiency results are provided in Appendix~\ref{app:efficiency}. Potential incorporates
connectivity, design-rule, wirelength, and via costs. These complementary
metrics distinguish a complete and rule-clean board from partial connectivity
or a route that is physically expensive.

% Declare the D2 float early so it can occupy the top of page 10.
\begin{table}[!t]
\centering
\captionsetup{justification=raggedright,singlelinecheck=false,skip=6pt}
\setlength{\tabcolsep}{4pt}
\renewcommand{\arraystretch}{1.08}
\caption{D2 routing quality (128 synthetic boards).}
\label{tab:routeagent-d2}
\small
\begin{tabular*}{\linewidth}{@{\extracolsep{\fill}}lrrrrrr@{}}
\toprule
Method & CP $\uparrow$ & Pot. $\uparrow$ & Rout. $\uparrow$ & DRV $\downarrow$ & WL (mm) $\downarrow$ & Via $\downarrow$ \\
\midrule
Freerouting & \textbf{1.000} & 14.42 & \textbf{1.000} & \textbf{0.00} & 409.8 & 2.48 \\
OrthoRoute & 0.010 & 0.38 & 0.340 & 8.54 & 323.4 & 11.30 \\
KiCadRoutingTools & \textbf{1.000} & 15.52 & \textbf{1.000} & \textbf{0.00} & 373.9 & 8.09 \\
\midrule
PCBWorld (Haiku 5.5) & 0.633 & 13.47 & 0.897 & 1.12 & 338.8 & 2.44 \\
PCBWorld (GPT-5.4) & 0.880 & 15.24 & 0.980 & 0.27 & 411.8 & 2.00 \\
PCBWorld (GPT-5.4-mini) & 0.200 & 6.46 & 0.490 & 4.62 & \textbf{208.6} & \textbf{0.96} \\
PCBWorld (GPT-5.4-nano) & 0.240 & 6.97 & 0.540 & 4.21 & 233.3 & 1.25 \\
PCBWorld (Qwen3.5-397B) & 0.140 & 5.81 & 0.480 & 4.64 & \textit{229.6} & \textit{1.02} \\
\midrule
PCBWorld (PPO) & \textit{0.980} & \textit{15.84} & \textit{0.990} & 0.06 & 458.4 & 2.22 \\
PCBWorld (GRPO) & 0.950 & 11.43 & 0.980 & 0.19 & 744.3 & 9.84 \\
PCBWorld (PPO terminal) & 0.610 & 12.06 & 0.800 & 1.60 & 380.2 & 1.62 \\
PCBWorld (PPO w/o finish) & \textbf{1.000} & 15.81 & \textbf{1.000} & \textit{0.01} & 497.1 & 1.93 \\
\midrule
PCBRouteAgent (Claude Haiku 5.5) & \textbf{1.000} & \textbf{16.27} & \textbf{1.000} & \textbf{0.00} & 392.4 & 1.99 \\
PCBRouteAgent (Claude Opus 5.5) & \textbf{1.000} & \textbf{16.27} & \textbf{1.000} & \textbf{0.00} & 392.5 & 1.99 \\
\bottomrule
\end{tabular*}
\end{table}

\paragraph{Comparisons.}
We place PCBRouteAgent alongside the methods in PCBWorld's~\cite{song2026pcbworld} main result table: the reference routes and the rule-based Freerouting
~\cite{freerouting2026}, OrthoRoute~\cite{benchoff2025orthoroute}, and
KiCadRoutingTools~\cite{haas2026kicadroutingtools} routers.
Interactive PCBWorld agents use GPT-5.4~\cite{openai2026gpt54}, GPT-5.4-mini and
GPT-5.4-nano~\cite{openai2026gpt54mini}, and
Qwen3.5-397B~\cite{qwenteam2026qwen35}. RL baselines use policies trained with
PPO~\cite{schulman2017ppo} or GRPO~\cite{shao2024deepseekmath}.

We evaluate
PCBRouteAgent with Claude Opus 5.5~\cite{anthropic2026opus55} and
Claude Haiku 5.5~\cite{anthropic2026haiku55} on all four splits. We also
locally evaluate the PCBWorld interactive agent with Haiku 5.5. The prompt template for the locally evaluated PCBWorld interactive baseline is listed in Appendix~\ref{app:prompt-pcbworld}.

To isolate the contribution of LLM-based guide revisions, we additionally
evaluate a w/o LLM variant on all four splits
(Table~\ref{tab:routeagent-planner-ablation}). We disable the model provider
in the final implementation. MST-based guide initialization, detailed
execution, and scripted local recovery remain enabled; LLM-based guide
revisions are omitted. All variants are scored
with the same official board evaluator.

Tables~\ref{tab:routeagent-d2}--\ref{tab:routeagent-d3b} combine the published
PCBWorld Table~22 (@1) baselines with our local evaluations. The
PCBWorld--Haiku rows are local reproductions. PCBWorld defines D3-C but
publishes no scores for it, so Table~\ref{tab:routeagent-d3c} contains our local
evaluations. The D3-B
PCBRouteAgent result includes retries of unscored model-response failures
and angle repair, so it is not a strict first-attempt @1 estimate.

\paragraph{Implementation details.}
We initialize routing corridors with an expansion margin of $\delta=1.5$ mm.
The controller uses a per-board replanning budget of 12, with at most four
replans per connection and five batch-replanning rounds. Local recovery
allows two stage backtracks, one same-layer A* detour, and two
alternative-anchor attempts per connection, with at most one rip-up per
obstructing net. For Claude Opus 5.5, D2 and D3-A use a maximum output length
of 32,000 tokens with medium reasoning effort; D3-B uses a 32,000-token output
limit, and D3-C uses 64,000 tokens with low effort. Reported Opus outputs
undergo up to five guarded angle-repair passes, accepting changes only when
connectivity, design-rule checks, and wirelength do not worsen. We score all saved boards with PCBWorld's official evaluation settings. The potential penalizes incomplete connections, error-level design-rule violations, wirelength, and vias; warning-level issues do not affect the score. Prompt templates are provided in Appendix~\ref{app:prompts}.

% Keep Results paragraphs compact without changing other sections.
\begingroup
\setlength{\parskip}{3pt}
\titlespacing*{\paragraph}{0pt}{6pt}{0.6em}
\section{Results and Analysis}
Tables~\ref{tab:routeagent-d2}--\ref{tab:routeagent-d3b} combine the published PCBWorld~\cite{song2026pcbworld} Table~22 (@1) baselines with our local evaluations. PCBWorld--Haiku is locally reproduced on each split, and PCBRouteAgent contributes one scored board per instance. Table~\ref{tab:routeagent-d3c} reports local evaluations on the fixed D3-C subset. Bold and italic values indicate the best and second-best distinct reported values, respectively, including Reference; ties share the same formatting.

\paragraph{Routing quality on synthetic boards.}
On D2 (Table~\ref{tab:routeagent-d2}), both PCBRouteAgent variants achieve a clean-pass rate of 1.00 and routability of 1.000, matching the complete routing achieved by Freerouting and KiCadRoutingTools. The Opus variant obtains a potential gain of 16.27, compared with 15.84 for PCBWorld PPO and 14.42 for Freerouting. Among these fully connected solutions, it reduces wirelength and via count relative to Freerouting by 4.2\% and 19.7\%, respectively. Thus, on the synthetic test set, PCBRouteAgent achieves complete, rule-clean routing with competitive physical routing costs.

\FloatBarrier

\begin{table}[!t]
\centering
\captionsetup{justification=raggedright,singlelinecheck=false,skip=6pt}
\setlength{\tabcolsep}{4pt}
\renewcommand{\arraystretch}{1.08}
\caption{D3-A routing quality (99 real-world boards).}
\label{tab:routeagent-d3a}
\small
\begin{tabular*}{\linewidth}{@{\extracolsep{\fill}}lrrrrrr@{}}
\toprule
Method & CP $\uparrow$ & Pot. $\uparrow$ & Rout. $\uparrow$ & DRV $\downarrow$ & WL (mm) $\downarrow$ & Via $\downarrow$ \\
\midrule
Reference & \textbf{1.000} & \textbf{23.16} & \textbf{1.000} & \textbf{0.00} & 170.2 & 1.18 \\
\midrule
Freerouting & 0.750 & 22.04 & 0.900 & 1.08 & 156.6 & 0.86 \\
OrthoRoute & 0.020 & -6.16 & 0.530 & 99.70 & 129.8 & 26.97 \\
KiCadRoutingTools & 0.740 & 20.00 & 0.940 & 2.38 & 142.8 & 3.69 \\
\midrule
PCBWorld (Haiku 5.5) & 0.343 & 14.77 & 0.721 & 4.06 & 111.5 & 0.96 \\
PCBWorld (GPT-5.4) & 0.420 & 16.15 & 0.810 & 2.71 & 148.1 & 2.36 \\
PCBWorld (GPT-5.4-mini) & 0.130 & 8.97 & 0.490 & 7.16 & \textbf{81.9} & \textbf{0.21} \\
PCBWorld (GPT-5.4-nano) & 0.150 & 8.35 & 0.470 & 7.46 & \textit{83.0} & \textit{0.31} \\
PCBWorld (Qwen3.5-397B) & 0.190 & 10.98 & 0.590 & 5.75 & 102.7 & 0.47 \\
\midrule
PCBWorld (PPO) & 0.680 & 19.22 & 0.870 & 1.91 & 164.9 & 1.09 \\
PCBWorld (GRPO) & 0.640 & 15.07 & 0.900 & 1.67 & 337.1 & 14.13 \\
PCBWorld (PPO terminal) & 0.650 & 18.23 & 0.860 & 1.95 & 171.0 & 3.50 \\
PCBWorld (PPO w/o finish) & 0.810 & 19.55 & 0.940 & 0.95 & 262.7 & 3.87 \\
\midrule
PCBRouteAgent (Claude Haiku 5.5) & 0.929 & 22.86 & 0.990 & 0.14 & 164.0 & 0.78 \\
PCBRouteAgent (Claude Opus 5.5) & \textit{0.949} & \textit{23.02} & \textit{0.994} & \textit{0.09} & 165.5 & 0.79 \\
\bottomrule
\end{tabular*}
\end{table}

\paragraph{Routing completion on small real-world boards.}
On D3-A (Table~\ref{tab:routeagent-d3a}), the Opus variant achieves a clean-pass rate of 0.949, compared with 0.810 for the strongest evaluated baseline, PCBWorld PPO w/o finish, an improvement of 13.9 percentage points. Its routability reaches 0.994, with an average DRV count of 0.09.

\FloatBarrier

\begin{table}[!t]
\centering
\captionsetup{justification=raggedright,singlelinecheck=false,skip=6pt}
\setlength{\tabcolsep}{4pt}
\renewcommand{\arraystretch}{1.08}
\caption{D3-B routing quality (10 real-world boards). Results include response-failure retries and angle repair.}
\label{tab:routeagent-d3b}
\small
\begin{tabular*}{\linewidth}{@{\extracolsep{\fill}}lrrrrrr@{}}
\toprule
Method & CP $\uparrow$ & Pot. $\uparrow$ & Rout. $\uparrow$ & DRV $\downarrow$ & WL (mm) $\downarrow$ & Via $\downarrow$ \\
\midrule
Reference & \textbf{1.000} & \textit{63.96} & \textbf{1.000} & \textbf{0.00} & 570.3 & 6.70 \\
\midrule
Freerouting & 0.640 & 58.86 & 0.990 & 11.94 & 533.5 & 3.88 \\
OrthoRoute & 0.000 & -10.25 & 0.440 & 188.20 & 370.0 & 39.80 \\
KiCadRoutingTools & 0.200 & 40.94 & 0.860 & 44.70 & 472.3 & 11.80 \\
\midrule
PCBWorld (Haiku 5.5) & 0.000 & 25.32 & 0.442 & 25.40 & \textit{245.7} & 1.90 \\
PCBWorld (GPT-5.4) & 0.000 & 25.65 & 0.440 & 24.88 & \textbf{240.3} & 2.06 \\
PCBWorld (GPT-5.4-mini) & 0.000 & 23.96 & 0.490 & 23.00 & 408.6 & \textit{0.94} \\
PCBWorld (GPT-5.4-nano) & 0.000 & 23.62 & 0.470 & 24.00 & 363.9 & \textbf{0.36} \\
PCBWorld (Qwen3.5-397B) & 0.000 & 19.02 & 0.410 & 26.84 & 249.6 & 6.24 \\
\midrule
PCBWorld (PPO) & 0.210 & 37.60 & 0.650 & 17.33 & 518.8 & 5.06 \\
PCBWorld (GRPO) & 0.070 & 21.76 & 0.550 & 22.28 & 1054.5 & 27.26 \\
PCBWorld (PPO terminal) & 0.150 & 33.47 & 0.640 & 17.54 & 550.0 & 10.64 \\
PCBWorld (PPO w/o finish) & 0.220 & 35.99 & 0.720 & 15.32 & 1008.6 & 14.57 \\
\midrule
PCBRouteAgent (Claude Haiku 5.5) & \textit{0.800} & 63.20 & \textit{0.997} & \textit{0.20} & 580.0 & 4.90 \\
PCBRouteAgent (Claude Opus 5.5) & \textbf{1.000} & \textbf{64.47} & \textbf{1.000} & \textbf{0.00} & 591.6 & 4.00 \\
\bottomrule
\end{tabular*}
\par\vspace{12pt}
\captionsetup{justification=raggedright,singlelinecheck=false,skip=6pt}
\setlength{\tabcolsep}{4pt}
\renewcommand{\arraystretch}{1.08}
\caption{D3-C routing quality (10 real-world boards); all method results are local evaluations.}
\label{tab:routeagent-d3c}
\small
\begin{tabular*}{\linewidth}{@{\extracolsep{\fill}}lrrrrrr@{}}
\toprule
Method & CP $\uparrow$ & Pot. $\uparrow$ & Rout. $\uparrow$ & DRV $\downarrow$ & WL (mm) $\downarrow$ & Via $\downarrow$ \\
\midrule
Reference & \textbf{0.900} & \textbf{147.80} & \textit{0.938} & \textit{8.80} & 1391.0 & 44.00 \\
\midrule
Freerouting & 0.300 & 132.80 & 0.918 & 32.00 & 1304.0 & 21.40 \\
OrthoRoute & 0.000 & 0.60 & 0.063 & 139.90 & \textbf{80.0} & 17.80 \\
KiCadRoutingTools & 0.000 & 115.60 & 0.778 & 30.50 & 1086.0 & 57.40 \\
\midrule
PCBWorld (Haiku 5.5) & 0.000 & 31.00 & 0.199 & 94.60 & \textit{330.9} & \textit{4.30} \\
\midrule
PCBWorld (PPO) & 0.000 & 52.20 & 0.403 & 72.70 & 1075.0 & 5.00 \\
PCBWorld (GRPO) & 0.000 & 44.80 & 0.353 & 79.00 & 886.0 & 8.30 \\
PCBWorld (PPO terminal) & 0.000 & 56.20 & 0.412 & 69.50 & 1072.0 & \textbf{1.50} \\
\midrule
PCBRouteAgent (Claude Haiku 5.5) & 0.100 & 135.99 & 0.915 & 9.80 & 1352.4 & 17.30 \\
PCBRouteAgent (Claude Opus 5.5) & \textit{0.400} & \textit{141.80} & \textbf{0.949} & \textbf{6.30} & 1410.5 & 21.30 \\
\bottomrule
\end{tabular*}
\end{table}

\paragraph{Routing completion on medium real-world boards.}
On D3-B (Table~\ref{tab:routeagent-d3b}), the Opus variant completes all ten boards without error-level violations, compared with a clean-pass rate of 0.64 for Freerouting.

\paragraph{Large-board routing remains challenging.}
On D3-C (Table~\ref{tab:routeagent-d3c}), the Opus variant achieves a clean-pass rate of 0.40, compared with 0.30 for Freerouting and 0.00 for the evaluated RL baselines and PCBWorld--Haiku agent. It reaches 0.949 mean routability and reduces the average DRV count from Freerouting's 32.0 to 6.3, an 80.3\% reduction. Nevertheless, six of the ten boards remain incomplete. The gap between high average routability and lower clean-pass rate shows that connecting most targets does not ensure successful completion of an entire board.

\FloatBarrier

\paragraph{Framework effectiveness with the same model.}
We do not conduct a full evaluation of PCBWorld's interactive agent with Claude Opus 5.5 because we verified that its per-step model invocation incurs substantially higher API costs than PCBRouteAgent. Instead, we compare the two frameworks using the same Claude Haiku 5.5 backbone (Tables~\ref{tab:routeagent-d2}--\ref{tab:routeagent-d3c}). PCBRouteAgent achieves clean-pass rates of 1.00, 0.929, 0.80, and 0.10 on D2, D3-A, D3-B, and D3-C, respectively, compared with 0.633, 0.343, 0.00, and 0.00 for PCBWorld--Haiku. These correspond to improvements of 36.7, 58.6, 80, and 10 percentage points. Mean routability also increases from 0.897 to 1.000 on D2, from 0.721 to 0.990 on D3-A, from 0.442 to 0.997 on D3-B, and from 0.199 to 0.915 on D3-C. The same language model therefore achieves higher connectivity and board completion within PCBRouteAgent, providing evidence for the effectiveness of the overall planning--execution framework.

\begin{table}[t]
\centering
\captionsetup{justification=raggedright,singlelinecheck=false,skip=6pt}
\setlength{\tabcolsep}{3pt}
\renewcommand{\arraystretch}{1.08}
\caption{Planner LLM ablation with the same detailed executor and scripted
local recovery. Bold and italic indicate the best and second-best distinct
values within each split; ties share formatting. Lower wirelength and via
count should be interpreted alongside routing completion.}
\label{tab:routeagent-planner-ablation}
\small
\begin{tabular*}{\linewidth}{@{\extracolsep{\fill}}llrrrrrr@{}}
\toprule
Set & Planner LLM & CP $\uparrow$ & Pot. $\uparrow$ & Rout. $\uparrow$ & DRV $\downarrow$ & WL (mm) $\downarrow$ & Via $\downarrow$ \\
\midrule
D2 (128) & w/o LLM & \textit{0.961} & \textit{16.09} & \textit{0.995} & \textit{0.04} & \textbf{390.7} & \textbf{1.95} \\
& Claude Haiku 5.5 & \textbf{1.000} & \textbf{16.27} & \textbf{1.000} & \textbf{0.00} & \textit{392.4} & \textit{1.99} \\
& Claude Opus 5.5 & \textbf{1.000} & \textbf{16.27} & \textbf{1.000} & \textbf{0.00} & 392.5 & \textit{1.99} \\
\midrule
D3-A (99) & w/o LLM & 0.535 & 20.13 & 0.922 & 1.00 & \textbf{150.8} & \textbf{0.22} \\
& Claude Haiku 5.5 & \textit{0.929} & \textit{22.86} & \textit{0.990} & \textit{0.14} & \textit{164.0} & \textit{0.78} \\
& Claude Opus 5.5 & \textbf{0.949} & \textbf{23.02} & \textbf{0.994} & \textbf{0.09} & 165.5 & 0.79 \\
\midrule
D3-B (10) & w/o LLM & 0.000 & 52.61 & 0.891 & 5.10 & \textbf{508.4} & \textbf{0.90} \\
& Claude Haiku 5.5 & \textit{0.800} & \textit{63.20} & \textit{0.997} & \textit{0.20} & \textit{580.0} & 4.90 \\
& Claude Opus 5.5 & \textbf{1.000} & \textbf{64.47} & \textbf{1.000} & \textbf{0.00} & 591.6 & \textit{4.00} \\
\midrule
D3-C (10) & w/o LLM & 0.000 & 123.48 & 0.831 & 18.70 & \textbf{1124.9} & \textbf{5.90} \\
& Claude Haiku 5.5 & \textit{0.100} & \textit{135.99} & \textit{0.915} & \textit{9.80} & \textit{1352.4} & \textit{17.30} \\
& Claude Opus 5.5 & \textbf{0.400} & \textbf{141.80} & \textbf{0.949} & \textbf{6.30} & 1410.5 & 21.30 \\
\bottomrule
\end{tabular*}
\end{table}

\paragraph{Contribution of LLM-based replanning.}
Table~\ref{tab:routeagent-planner-ablation} compares the full framework
with its w/o LLM variant while retaining the same detailed executor and
scripted local recovery. Without LLM-based guide revisions, clean-pass
rates are 0.961, 0.535, 0.000, and 0.000 on D2, D3-A, D3-B, and D3-C,
respectively. Adding Opus raises these rates to 1.000, 0.949, 1.000, and
0.400, gains of 3.9, 41.4, 100, and 40 percentage points. Haiku also
improves clean-pass rates on all four splits. The distinction between
partial connectivity and board completion is particularly clear on D3-B
and D3-C: without the LLM, mean routability reaches 0.891 and 0.831,
yet no board passes cleanly. LLM-based revisions therefore contribute
to completing connections left unresolved by the scripted pipeline.
These gains come with increased wirelength and via count; for example,
on D3-A, Opus increases mean wirelength from 150.8 to 165.5 mm and via
count from 0.22 to 0.79. This ablation establishes the benefit relative
to disabling LLM-based revisions, but does not compare against a
dedicated rule-based replanner or a general-purpose agent using the
same execution tools.

\paragraph{Computational efficiency under the same conditions.}
The local Haiku comparison also shows that PCBRouteAgent requires fewer computational resources (Appendix~\ref{app:efficiency}). Across all 247 boards, model calls decrease from 22,529 to 172, combined input and output tokens from 105.7M to 12.0M, and API cost from \$20.72 to \$4.77. With the same timing scope, cumulative per-board runtime decreases from 36.4 to 2.6 hours. These correspond to approximately $131\times$ fewer model calls, $8.8\times$ lower token usage, $4.3\times$ lower API cost, and $14\times$ lower cumulative runtime. Together with the higher clean-pass rates on all four splits, these results show that the planning--execution framework improves both routing completion and computational efficiency.

\paragraph{Completion and physical routing cost.}
Improved completion need not produce shorter routing. On D3-B, the Opus variant uses 10.9\% more wirelength than Freerouting while improving clean-pass rate by 36 percentage points. On D3-C, its wirelength is 8.2\% higher, with a comparable via count of 21.3 versus 21.4. These results indicate a remaining opportunity to improve routing compactness while maintaining connectivity. Wirelength and via count must also be interpreted alongside completion: methods that leave many connections unrouted can obtain low physical costs simply by constructing less copper.

\paragraph{Remaining gap to PCB engineering.}
Our case studies reveal two limitations beyond aggregate routing scores (Appendix~\ref{app:case-studies}). First, early routing decisions can restrict access to pads and prevent later connections from being completed despite local recovery. Second, complete connectivity does not ensure that signal-specific requirements, such as differential-pair routing~\cite{siliconlabs2020usb}, are addressed. In the USB example (Figure~\ref{fig:routeagent-case-electrical}), PCBRouteAgent completes all 55 connections but routes D+ and D- independently, without explicitly representing their differential-pair relationship. This result calls for electrical validation beyond connectivity and basic design-rule checks. These findings indicate that approaching the capabilities of a PCB routing engineer requires stronger coordination across routing decisions and explicit modeling of signal-specific constraints.

\begin{figure}[!ht]
\centering
\includegraphics[width=\linewidth]{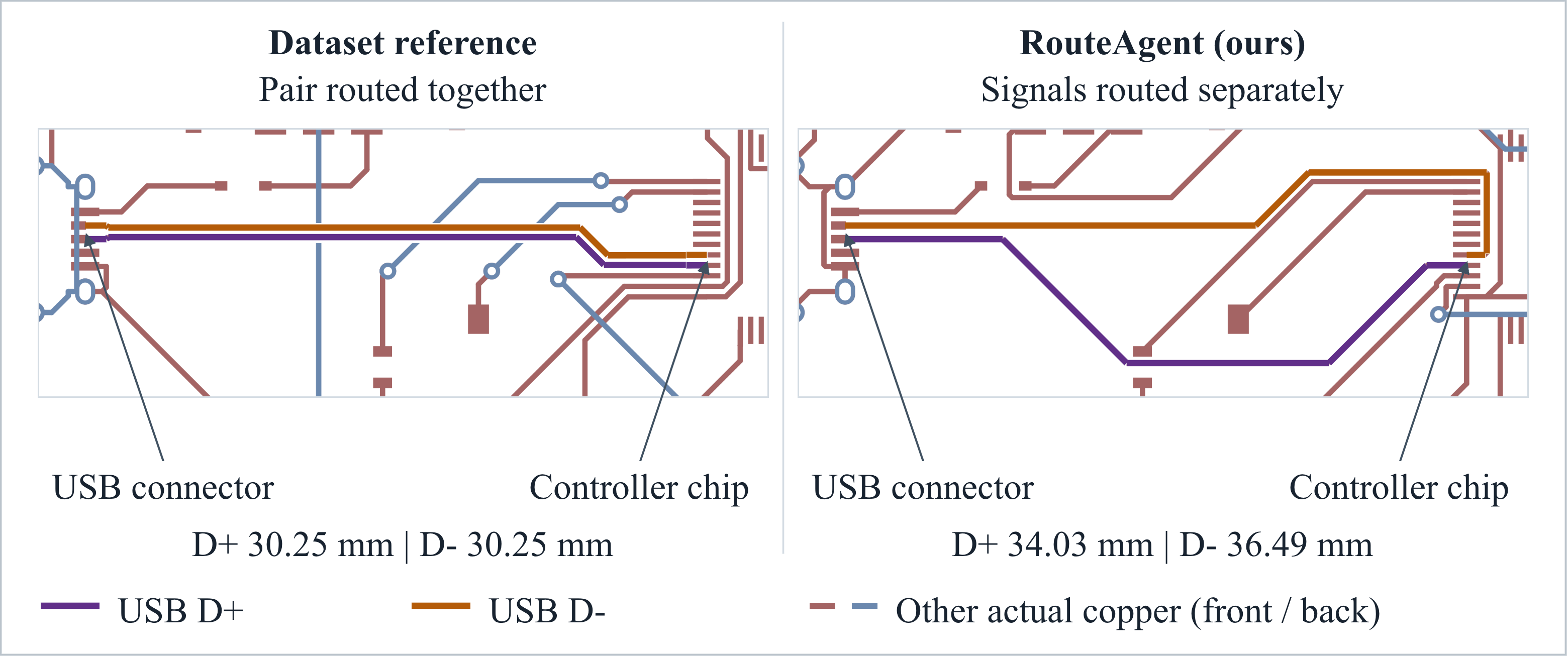}
\caption{USB D+ and D- in the reference (left) and the fully connected
PCBRouteAgent output (right). The reference routes the pair together;
PCBRouteAgent produces separate paths.}
\label{fig:routeagent-case-electrical}
\end{figure}

\FloatBarrier

\endgroup
\section{Conclusion}
We presented PCBRouteAgent, a multimodal PCB routing agent that coordinates global planning and native geometric execution through structured feedback. Inspired by the interaction between board-level strategy and local routing decisions in engineering workflows, the framework combines routing guides, bounded local recovery, and targeted guide revisions. Evaluation on 247 synthetic and real-world boards shows that PCBRouteAgent achieves complete, rule-clean routing on D2 and D3-B and improves clean-pass rate over the best evaluated baseline on all three real-world subsets. Large-board completion and routing compactness remain important challenges. Future work will investigate more effective spatial and layer planning, adaptive recovery budgets, and routing refinement that reduces physical cost while preserving completed connections.

\clearpage

\bibliographystyle{unsrt}
\bibliography{references}

\clearpage
\appendix
\section{Overall Routing Algorithm}
\label{app:routing-algorithm}
Algorithm~\ref{alg:pcbrouteagent} summarizes PCBRouteAgent. Initialization follows Sections~\ref{sec:global-agent}--\ref{sec:feedback-replanning}; local recovery follows Section~\ref{sec:local-recovery}. The LLM call uses Equation~\ref{eq:llm-replanning} and the prompt templates in Appendix~\ref{app:prompts}.

\begin{algorithm}[H]
\caption{PCBRouteAgent}
\label{alg:pcbrouteagent}
\begin{algorithmic}[1]
\footnotesize
\Require Initial PCB $B_0$, target nets $\mathcal{N}$, design rules $\mathcal{R}$, LLM $\mathcal{M}_{\theta}$, execution budget, maximum replanning rounds $r_{\max}$
\Ensure Saved PCB $B_r$ and execution report
\State $(\mathcal{I},S) \gets \Call{ExtractRepresentation}{B_0,\mathcal{N},\mathcal{R}}$
\State $\mathcal{G}_0 \gets \Call{InitializeGuide}{B_0,\mathcal{N},\mathcal{R}}$ \Comment{MST, order, corridors, via regions}
\State $B \gets B_0$; $t \gets 0$
\While{execution budget remains}
    \State $\mathcal{T} \gets \Call{CompilePendingTasks}{\mathcal{G}_t,B}$
    \If{$\mathcal{T}=\varnothing$}
        \State \textbf{break}
    \EndIf
    \State $\mathcal{D}_t \gets \varnothing$
    \For{each connection task $e$ in $\mathcal{T}$, in planned order}
        \State $C \gets \Call{Checkpoint}{B}$
        \If{checkpoint creation fails}
            \State \textbf{go to} finalization
        \EndIf
        \State $(B',\mathcal{F}_e) \gets \Call{ExecuteAndRecover}{e,B,\mathcal{R}}$
        \If{connectivity and board-state checks pass}
            \State $B \gets B'$ \Comment{Accept the connection}
        \Else
            \State $B \gets \Call{Restore}{C}$; $\mathcal{D}_t \gets \mathcal{D}_t\cup\{\mathcal{F}_e\}$
        \EndIf
        \If{execution budget is exhausted or rollback fails}
            \State \textbf{go to} finalization
        \EndIf
    \EndFor
    \If{all target connections are established or $\mathcal{D}_t=\varnothing$ or $t\geq r_{\max}$}
        \State \textbf{break}
    \EndIf
    \State $\mathcal{K}_t \gets \Call{ConnectivitySummaries}{B,\mathcal{N}}$
    \State $\Delta\mathcal{G}_t \gets \mathcal{M}_{\theta}(\mathcal{I},S,\mathcal{G}_t,\mathcal{D}_t,\mathcal{K}_t;P_{\mathrm{sys}},P_{\mathrm{replan}})$
    \State Validate $\Delta\mathcal{G}_t$; request a bounded correction if needed
    \State $\mathcal{G}_{t+1} \gets \mathcal{G}_t$
    \If{$\Delta\mathcal{G}_t$ is valid}
        \State $\mathcal{G}_{t+1} \gets \Call{ApplyPatch}{\mathcal{G}_t,\Delta\mathcal{G}_t}$ \Comment{Preserve accepted routes}
    \EndIf
    \State $t \gets t+1$
\EndWhile
\State \textbf{Finalization:} $B_r \gets \Call{FinalizeAndSave}{B}$ \Comment{Last consistent board state}
\State Evaluate $B_r$ using the official evaluator
\State \Return $B_r$ and execution report
\end{algorithmic}
\end{algorithm}

\clearpage
\section{Prompt Templates}
\label{app:prompts}
{\raggedright
The following templates reproduce the fixed prompt wording used by the implementation. Brace-delimited placeholders denote runtime-expanded board data, view metadata, feedback, routing guides, JSON schemas, or examples. The Global planner templates are extracted from \path{route_agent/prompts/bundle.py}; the correction fields follow \path{route_agent/cli/pipeline.py}. The locally evaluated PCBWorld baseline uses the v5 templates in \path{methods/llm_agent/wrappers/prompts/v5.py} with S-expression board observations. Input images are attached separately to the Global planner requests.\par}

\subsection{Global Routing Planner Prompts}

\subsubsection{System Prompt}
\label{app:prompt-global-system}
\begin{promptbox}{System Prompt for the Global Routing Planner}
\raggedright
\promptfield{Prompt}\par\smallskip
% Extracted from implementation; dynamic payloads use brace placeholders.
You are the Global routing planner for a printed circuit board.\par
You receive (1) board-context, per-layer copper-state and connectivity-demand images of the\par
current committed board in one physical mm frame and (2) a\par
structured sidecar with exact ids and rules. You output a RouteGuidePlan as JSON only.\par
\smallskip
What a plan is:\par
- connection\_trees: per net, edges that join two EXISTING connectivity clusters\par
  (ids like cluster:24:0). Never invent a terminal; Steiner regions are corridors, not clusters.\par
- corridors: axis-aligned regions per layer (bbox [x1,y1,x2,y2] in mm). They are SOFT: the\par
  Detailed router works inside them when it can and reports deviations. A region may cover\par
  obstacles; the Detailed router finds a legal path inside it.\par
- An edge lists its ordered stages (corridor ids). Consecutive same-layer stages must\par
  overlap. A layer change needs the two stages to overlap AND a via\_region covering that\par
  overlap, listed in via\_regions in order.\par
- Layer changes are YOUR decision only: the Detailed router never places a via and never\par
  leaves the layer of the stage it is in. A pad fenced in on its own layer (a dense SMD row,\par
  a pad walled by neighbouring pads or earlier tracks) cannot be left on that layer: start\par
  the edge on the other layer when the pad is thru-hole, otherwise put a small via\_region\par
  (about 1 mm) right next to the pad between a short first stage on the pad's layer and the\par
  next stage on the other layer. Without that the edge is certain to fail.\par
- The first stage must contain an anchor of the source cluster on that layer; the last\par
  stage must contain an anchor of the target cluster.\par
- profile\_ref must be one of the allowed profiles.\par
- net\_order is your routing order; reservations are soft preferences.\par
- based\_on.snapshot\_id must copy the sidecar's snapshot\_id exactly.\par
\smallskip
Plan only the work-set nets. Prefer short, wide corridors; use layer changes where a pad or\par
a channel needs one and nowhere else; keep corridors inside the board bbox; use the images\par
to avoid dense pad fields and existing copper. Do not draw copper. Output JSON matching the\par
schema, nothing else.\par

\end{promptbox}

\subsubsection{Feedback-Driven Replanning Prompt}
\label{app:prompt-replanning}
\begin{promptbox}{Prompt for Feedback-Driven Replanning}
\raggedright
\promptfield{Images} Board-context, layer-specific copper-state, and connectivity-demand views.\par\smallskip
\promptfield{Prompt}\par\smallskip
% Extracted from implementation; dynamic payloads use brace placeholders.
\# Task\par
Produce a RouteGuidePlan for the work-set nets of this board.\par
\smallskip
\# Images\par
One frame, mm coordinates; pixel\_to\_mm affine per view is in the sidecar.views list.\par
- \{view\_id\}: layers \{view\_layers\}, bbox\_mm \{view\_bbox\_mm\}, size \{view\_image\_size\}\par
\smallskip
\# Structured sidecar (JSON)\par
\{structured\_board\_json\}\par
\smallskip
\# Previous plan (revise only the affected items; keep the rest)\par
\{current\_routing\_guide\_json\}\par
\smallskip
\# Detailed feedback (the failure you must fix)\par
\{feedback\_batch\_and\_live\_state\_json\}\par
\smallskip
Return a GlobalGuidePatch JSON object: \{patch\_id, based\_on: \{snapshot\_id, rules\_hash, plan\_revision: \{plan\_revision\}\}, replace\_corridors, replace\_via\_regions, replace\_edges, preserve, net\_order?, budget\_revision?\}. Replace only what the feedback requires (a replaced edge or corridor is given in full); list every unaffected edge id in preserve. The cluster ids and snapshot\_id stay those of the sidecar (the board was rolled back to that stable state plus the edges already done, see feedback.live\_state). A full RouteGuidePlan is accepted instead of a patch but must then validate on its own.\par
\smallskip
\# GlobalGuidePatch JSON schema\par
\{patch\_json\_schema\}\par
\smallskip
\# Example patch (shape only; use the real ids, layers and mm coordinates of this board)\par
\{patch\_example\_json\}\par
\smallskip
If the feedback carries `packets` (several edges, an end-of-pass batch), every edge in it failed in this pass and the board now holds everything else that was routed: return ONE patch that fixes as many of them as you can, those whose evidence makes the fix clear first (via\_sites, head\_is\_thru, alt\_starts), and list every edge you leave unchanged in preserve.\par
\smallskip
The Detailed never places a via on its own: a layer change exists only where the plan has a via region. When failure\_evidence.via\_sites is non-empty the head is fenced in on its layer; each entry is a clearance-checked [x, y] in mm right next to the stuck head where a via fits on both layers. To route the edge on the other layer, add a via region (a box about 1 mm wide around one of those points), a short first-stage corridor on the head's layer that contains the head and overlaps that box, and the next stage on the other layer overlapping it too. blocker\_candidates lists the foreign pads / tracks fencing the head in (with positions and layers); head\_is\_thru true means the pad itself spans both layers, so a corridor starting on the other layer needs no via; alt\_starts are other points of the net's copper near the first stage the edge could start from instead.\par
\smallskip
When a packet carries `history`, the edge has been re-planned before: each entry is the shape the Detailed executed in an earlier round (stages with layer and bbox, via regions) and where it died (died\_at, head, reason). Those shapes have already failed: do not return one of them or a near copy. Change what failed: enter the target from its other side or on the other layer, move the via region away from died\_at, or start the edge from one of alt\_starts.\par
\smallskip
Rules the validator enforces: every corridor has id, layer and bbox; every via region has id, layers (the two layers it joins) and bbox; consecutive same-layer stages overlap; a layer change needs a via region covering the overlap of the two stages; the first stage contains an anchor of source\_cluster on its layer and the last stage an anchor of target\_cluster on its layer.\par

\smallskip
\promptfield{Output Format} A JSON \texttt{GlobalGuidePatch}; the serialized schema and example are inserted at the indicated placeholders.\par
\end{promptbox}

\subsection{Guide-Patch Correction Prompt}
\label{app:prompt-repair}
The correction request reuses the system prompt and replanning user template above. Its feedback object additionally contains the following fields, including the exact correction instruction. The previous response and validation-error list are populated at runtime.
\begin{promptbox}{Additional Feedback Fields for Guide-Patch Correction}
\raggedright
\promptfield{Prompt}\par\smallskip
% Extracted from implementation; dynamic payloads use brace placeholders.
"previous\_response": \{previous\_response\_json\},\par
"validation\_errors": \{validation\_errors\_json\},\par
"instruction": "Your previous patch failed the checks listed in validation\_errors. Return a corrected GlobalGuidePatch for the same edge(s); keep everything else unchanged."\par

\smallskip
\promptfield{Output Format} A corrected JSON \texttt{GlobalGuidePatch}.\par
\end{promptbox}

\subsection{PCBWorld Interactive Baseline Prompt}
\label{app:prompt-pcbworld}
The system message contains the static board context; the user message supplies the current observation, action history, rejected-attempt information, and valid actions at each step.
\begin{promptbox}{System Prompt for the PCBWorld Interactive Baseline}
\raggedright
\promptfield{Prompt}\par\smallskip
% Extracted from implementation; dynamic payloads use brace placeholders.
You are a PCB routing agent for KiCad. Connect unconnected pins (points in routing\_geometry) within each net. Never connect pins from different nets.\par
\smallskip
\# Priority\par
1. Complete all connections (reduce unconnected points to zero).\par
2. Avoid DRC violations (no crossing other nets, respect clearance).\par
3. Minimize total wirelength.\par
\smallskip
\# Routing Guidelines\par
- Targets are the (point ...) entries in routing\_geometry; the final segment must land on the target pad's (x, y) while the cursor is on that pad's layer (points carry no layer — look up the pad in board\_static or the nearby tracks).\par
- Use exact coordinates from the observation (pad positions, point targets, track endpoints). Only invent coordinates for detour waypoints.\par
- Avoid static obstacles, board edges, and existing tracks on the same layer (any net, including the current one).\par
- When a path is blocked, detour: switch layers with `make\_via` (offset \(\geq\)1mm from any pad — never place a via on a pad), route around on the opposite layer, then switch back.\par
- When the cursor is already near the remaining target with a clear path, prefer `finish <mode>` over a manual `make\_line`.\par
\smallskip
\#\# Layer-change detour pattern\par
Route from (0, 0) on layer 1 to (10, 10) on layer 1, bypassing an obstacle via layer 2:\par
  1. start\_route 0.0 0.0 1\par
  2. make\_via 0.0 0.5 w         \# offset off the source pad, then cross to layer 2\par
  3. start\_route 0.0 0.5 2      \# re-enter routing on the new layer\par
  4. make\_via 10.0 9.5 w        \# STOP before the target's (x, y) and cross back\par
  5. start\_route 10.0 9.5 1     \# re-enter routing on the target pad's layer\par
  6. make\_line 10.0 10.0 w      \# final segment lands on the pad, on its own layer\par
\smallskip
\#\# Do NOT draw or end on a pad from the wrong layer\par
Never let `make\_line` or `make\_via` reach the target pad's exact (x, y) while the cursor is on a different layer than that pad. If you do, you are trapped:\par
  - placing a via at the pad violates the "no via on pad" rule, and\par
  - the pad is unreachable from the wrong layer, so `finish` will fail repeatedly.\par
Always make the layer-return `make\_via` at an offset waypoint (\(\geq\)1mm away from the target pad) *before* the track arrives at the target's (x, y).\par
\smallskip
\# Response Format\par
Output exactly one <think>...</think> followed by one <action>...</action>. Never repeat think-action pairs.\par
"[no effect]" in history = action made no progress; "Last rejected attempt" below = action was refused. Do not repeat either.\par
\smallskip
\# Board State\par
The board state is provided in S-expression format. All floating-point values are rounded to 3 decimal places. Each element is enclosed in parentheses: (tag value ...) or (tag (child ...) ...).\par
\smallskip
\#\# board\_static — static board context (unchanged during episode)\par
  (bbox x y w h)          — board bounding box origin and size in mm\par
  (scale N)               — max(w, h), for normalization reference\par
  (net\_count N)           — total number of nets\par
  (copper\_layers N)       — number of copper layers\par
  (boardlines (edge ID (p1 PID x y) (p2 PID x y)) ...)  — board outline edges; an arc outline entry appears as (arc ID (p1 ...) (mid PID x y) (p2 ...)) with mid on the arc between p1 and p2 (full circle: p1 == p2, mid = opposite point)\par
  (nets (net CODE "NAME" (pads (pad ID x y layer) ...)) ...)   — nets with pad center positions; layer is a copper layer number (1=Top, N=Bottom) or `th` for a through-hole pad (plated through every copper layer — start\_route / make\_via may land on any layer)\par
  (obstacles (obs ID x y (size w h) (layer N)) ...)      — static obstacles\par
\smallskip
\#\# routing\_geometry — dynamic per-net routing state (updated every step)\par
  (net CODE\par
    (tracks (track ID (p1 PID x y) (p2 PID x y) (layer N)) ...)  — placed track segments\par
    (vias (via ID (center PID x y) (layers start end)) ...)       — placed vias\par
    (points (point ID x y) ...))                                  — unconnected ratsnest targets; layer info not provided (infer from the target pad or nearby tracks)\par
\smallskip
\#\# router\_head — current router cursor state\par
  (xy x y)             — cursor position in mm\par
  (layer N)            — active copper layer (1=Top)\par
  (net N)              — currently selected net code (0=none)\par
  (phase N)            — 0=NET\_SELECT, 1=START\_ROUTE, 2=ROUTING\par
  (is\_routing true|false)\par
  (routing\_mode M)     — m=MarkObstacles, p=PushAndShove, w=Walkaround\par
  (step N ratio)       — current step number and step/max\_steps ratio\par
\smallskip
\{board\_static\}\par

\end{promptbox}

\begin{promptbox}{User Prompt for the PCBWorld Interactive Baseline}
\raggedright
\promptfield{Prompt}\par\smallskip
% Extracted from implementation; dynamic payloads use brace placeholders.
\#\# Step \{current\_step\}\par
\{dynamic\_observation\}\par
\smallskip
\# History (step \{current\_step\}, \{valid\_step\_count\} valid actions taken)\par
\{action\_history\}\{rejected\_attempts\}\par
\smallskip
\# Valid Actions\par
Choose exactly one from:\par
\{valid\_actions\}\par

\smallskip
\promptfield{Output Format} One \texttt{<think>...</think>} block followed by one \texttt{<action>...</action>} block, as specified in the system prompt.\par
\end{promptbox}

\clearpage
\section{Computational Cost and Efficiency}
\label{app:efficiency}

We compare PCBWorld's interactive agent and PCBRouteAgent using Claude Haiku 5.5 under the same local evaluation conditions. Table~\ref{tab:haiku-per-board-cost} reports per-board means for each test split. Time uses the same measurement scope for both methods. Model calls count requests to the language model, while input and output tokens and API charges are taken from the recorded usage. Clean-pass rates match the corresponding Haiku rows in the main routing-quality tables. All displayed values are rounded.

\begin{table}[!htbp]
\centering
\captionsetup{justification=raggedright,singlelinecheck=false,skip=6pt}
\setlength{\tabcolsep}{3pt}
\renewcommand{\arraystretch}{1.08}
\caption{Per-board routing success and computational cost with Claude Haiku 5.5 under the same evaluation conditions. Input and output tokens are in thousands (k). Within each split, bold and italic indicate the better and second-best values, respectively.}
\label{tab:haiku-per-board-cost}
\small
\begin{tabular*}{\linewidth}{@{\extracolsep{\fill}}llrrrrrr@{}}
\toprule
Set & Method & CP $\uparrow$ & Time (s) $\downarrow$ & Calls $\downarrow$ & In. (k) $\downarrow$ & Out. (k) $\downarrow$ & Cost (\$) $\downarrow$ \\
\midrule
D2 & PCBWorld & \textit{0.633} & \textit{188.0} & \textit{43.20} & \textit{45.8} & \textit{38.9} & \textit{0.0240} \\
(128) & PCBRouteAgent & \textbf{1.000} & \textbf{0.8} & \textbf{0.05} & \textbf{1.6} & \textbf{0.1} & \textbf{0.0002} \\
\midrule
D3-A & PCBWorld & \textit{0.343} & \textit{436.0} & \textit{76.60} & \textit{112.6} & \textit{88.1} & \textit{0.0553} \\
(99) & PCBRouteAgent & \textbf{0.929} & \textbf{27.0} & \textbf{0.87} & \textbf{19.9} & \textbf{5.8} & \textbf{0.0048} \\
\midrule
D3-B & PCBWorld & \textit{0.000} & \textit{1817.0} & \textit{263.90} & \textit{900.0} & \textit{352.0} & \textit{0.2660} \\
(10) & PCBRouteAgent & \textbf{0.800} & \textbf{146.0} & \textbf{2.90} & \textbf{163.0} & \textbf{30.0} & \textbf{0.0330} \\
\midrule
D3-C & PCBWorld & \textit{0.000} & \textit{4566.0} & \textit{676.60} & \textit{5430.0} & \textit{816.0} & \textit{0.9510} \\
(10) & PCBRouteAgent & \textbf{0.100} & \textbf{516.0} & \textbf{5.10} & \textbf{666.0} & \textbf{62.0} & \textbf{0.3930} \\
\bottomrule
\end{tabular*}
\end{table}

\paragraph{Efficiency relative to completed work.}
Table~\ref{tab:haiku-normalized-cost} relates resource use to completed connections and clean-pass boards. Tokens per connected edge divide the total input and output tokens by the number of connected target edges across the split. Cost per clean-pass board divides the total API cost of all evaluated boards, including failed boards, by the number of boards that pass. It is undefined when no board passes, denoted by a dash. This is an aggregate evaluation cost per success, not a measurement of repeated attempts on an individual board.

\begin{table}[!htbp]
\centering
\captionsetup{justification=raggedright,singlelinecheck=false,skip=6pt}
\setlength{\tabcolsep}{4pt}
\renewcommand{\arraystretch}{1.08}
\caption{Resource use normalized by completed work with Claude Haiku 5.5. Connected edges are reported as completed/required totals for each split. Lower token and API costs are better.}
\label{tab:haiku-normalized-cost}
\small
\begin{tabular*}{\linewidth}{@{\extracolsep{\fill}}llrrr@{}}
\toprule
Set & Method & Connected edges & Tokens/edge $\downarrow$ & \$/clean-pass board $\downarrow$ \\
\midrule
D2 & PCBWorld & \textit{953/1096} & \textit{11,374} & \textit{0.0380} \\
 & PCBRouteAgent & \textbf{1096/1096} & \textbf{202} & \textbf{0.0002} \\
\midrule
D3-A & PCBWorld & \textit{900/1301} & \textit{22,079} & \textit{0.1610} \\
 & PCBRouteAgent & \textbf{1287/1301} & \textbf{1,972} & \textbf{0.0050} \\
\midrule
D3-B & PCBWorld & \textit{191/445} & \textit{65,545} & --- \\
 & PCBRouteAgent & \textbf{443/445} & \textbf{4,355} & 0.0410 \\
\midrule
D3-C & PCBWorld & \textit{232/1178} & \textit{269,324} & --- \\
 & PCBRouteAgent & \textbf{1080/1178} & \textbf{6,744} & 3.9300 \\
\bottomrule
\end{tabular*}
\end{table}

\paragraph{Aggregate resource use.}
Table~\ref{tab:haiku-total-cost} summarizes usage across all 247 boards. The cumulative time is the sum of per-board elapsed times, not the elapsed duration of a parallel batch. PCBRouteAgent reduces model calls by approximately $131\times$, combined input and output tokens by $8.8\times$, API cost by $4.3\times$, and cumulative time by $14\times$. It also requires fewer tokens per connected edge on every split (Table~\ref{tab:haiku-normalized-cost}). Large-board routing remains costly relative to successful completion: on D3-C, only one of ten PCBRouteAgent boards passes, resulting in an aggregate API cost of \$3.93 per clean-pass board.

\begin{table}[!htbp]
\centering
\captionsetup{justification=raggedright,singlelinecheck=false,skip=6pt}
\setlength{\tabcolsep}{4pt}
\renewcommand{\arraystretch}{1.08}
\caption{Total computational cost across all 247 boards with Claude Haiku 5.5. Token counts are in millions (M); cumulative time sums the elapsed time of individual boards. Lower is better in every column.}
\label{tab:haiku-total-cost}
\small
\begin{tabular*}{\linewidth}{@{\extracolsep{\fill}}lrrrrr@{}}
\toprule
Method & Calls & In. tokens (M) & Out. tokens (M) & API cost (\$) & Time (h) \\
\midrule
PCBWorld & \textit{22,529} & \textit{80.3} & \textit{25.4} & \textit{20.72} & \textit{36.4} \\
PCBRouteAgent & \textbf{172} & \textbf{10.5} & \textbf{1.5} & \textbf{4.77} & \textbf{2.6} \\
\bottomrule
\end{tabular*}
\end{table}
\FloatBarrier

% Appendix-only case studies; existing manuscript sections are unchanged.
% Figures retain the reviewed board geometry and labels.
\clearpage
\section{Case Studies and Failure Analysis}
\label{app:case-studies}

\subsection{Same-board comparison with PCBWorld}
\begin{figure}[!ht]
\centering
\includegraphics[width=\linewidth]{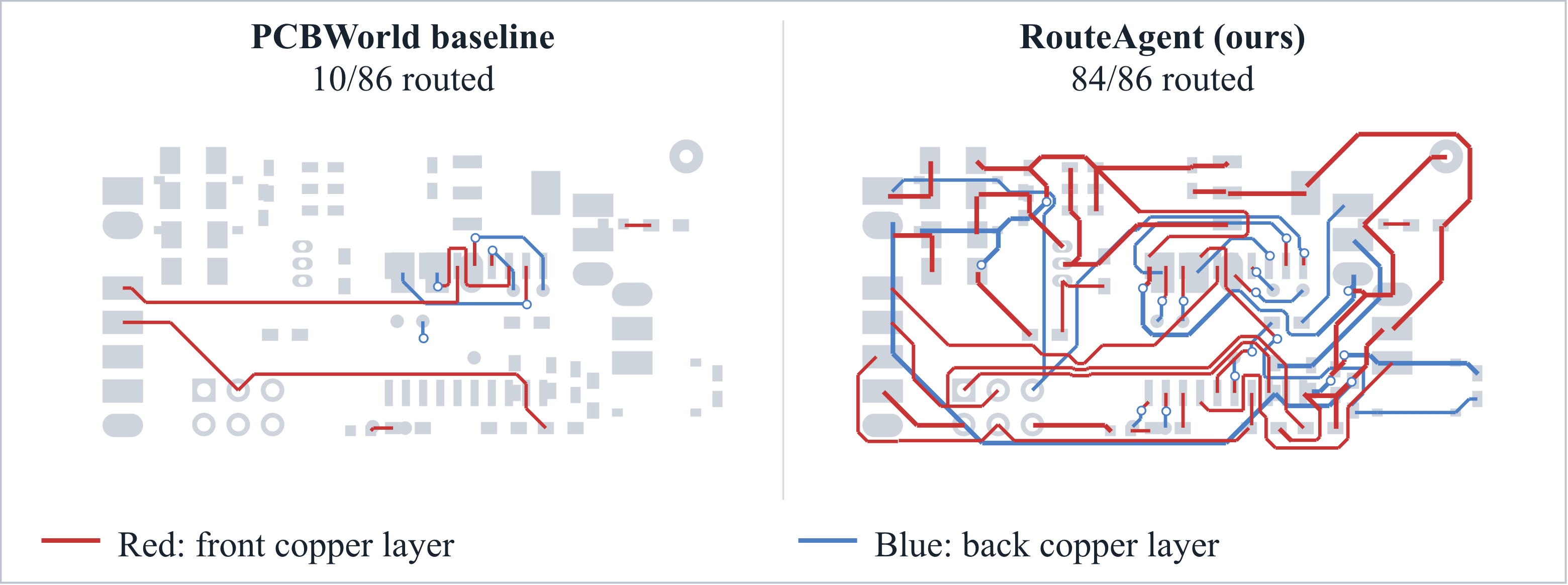}
\caption{Saved routing outputs for the same D3-C board with Claude
Haiku~5.5.}
\label{fig:routeagent-case-comparison}
\end{figure}

On a D3-C board with the same Claude Haiku~5.5 backbone,
PCBRouteAgent completes 84 of 86 connections, compared with 10 for
the PCBWorld baseline, an improvement of 86.0 percentage points
(Figure~\ref{fig:routeagent-case-comparison}).
Table~\ref{tab:routeagent-case-resources} shows that this improvement
also uses fewer model calls and tokens.

\begin{table}[!ht]
\centering
\captionsetup{justification=raggedright,singlelinecheck=false,skip=6pt}
\setlength{\tabcolsep}{3pt}
\renewcommand{\arraystretch}{1.08}
\caption{Same-board routing quality and resource use with Claude
Haiku~5.5.}
\label{tab:routeagent-case-resources}
\small
\begin{tabular*}{\linewidth}{@{\extracolsep{\fill}}lrrrrrrr@{}}
\toprule
Method & Pot. $\uparrow$ & Rout. $\uparrow$ &
\shortstack{WL\\(mm) $\downarrow$} & Via $\downarrow$ &
\shortstack{Model\\calls $\downarrow$} &
\shortstack{Input\\tokens $\downarrow$} &
\shortstack{Output\\tokens $\downarrow$} \\
\midrule
PCBWorld baseline & 13.72 & 0.116 & 113.29 & 7 & 403 & 2,024,617 & 682,784 \\
PCBRouteAgent (ours) & 107.73 & 0.977 & 657.16 & 19 & 5 & 398,021 & 86,963 \\
\bottomrule
\end{tabular*}
\end{table}
\FloatBarrier

The PCBWorld baseline uses one model call per action and reaches
the board's 403-step action limit. With five calls, PCBRouteAgent
reduces recorded input and output tokens by 80.3\% and 87.3\%, respectively.
This result is consistent with the planning--execution design:
the model plans groups of connections, while the routing engine
performs the individual path-search and copper-placement steps.
When progress stalls, the model can revise a group of connections
without selecting each low-level action separately. The comparison
therefore illustrates how the framework can use the same model's
decisions more efficiently while completing substantially more routing.

\FloatBarrier

\subsection{Failure analysis of incomplete routing}
\begin{figure}[!ht]
\centering
\includegraphics[width=\linewidth]{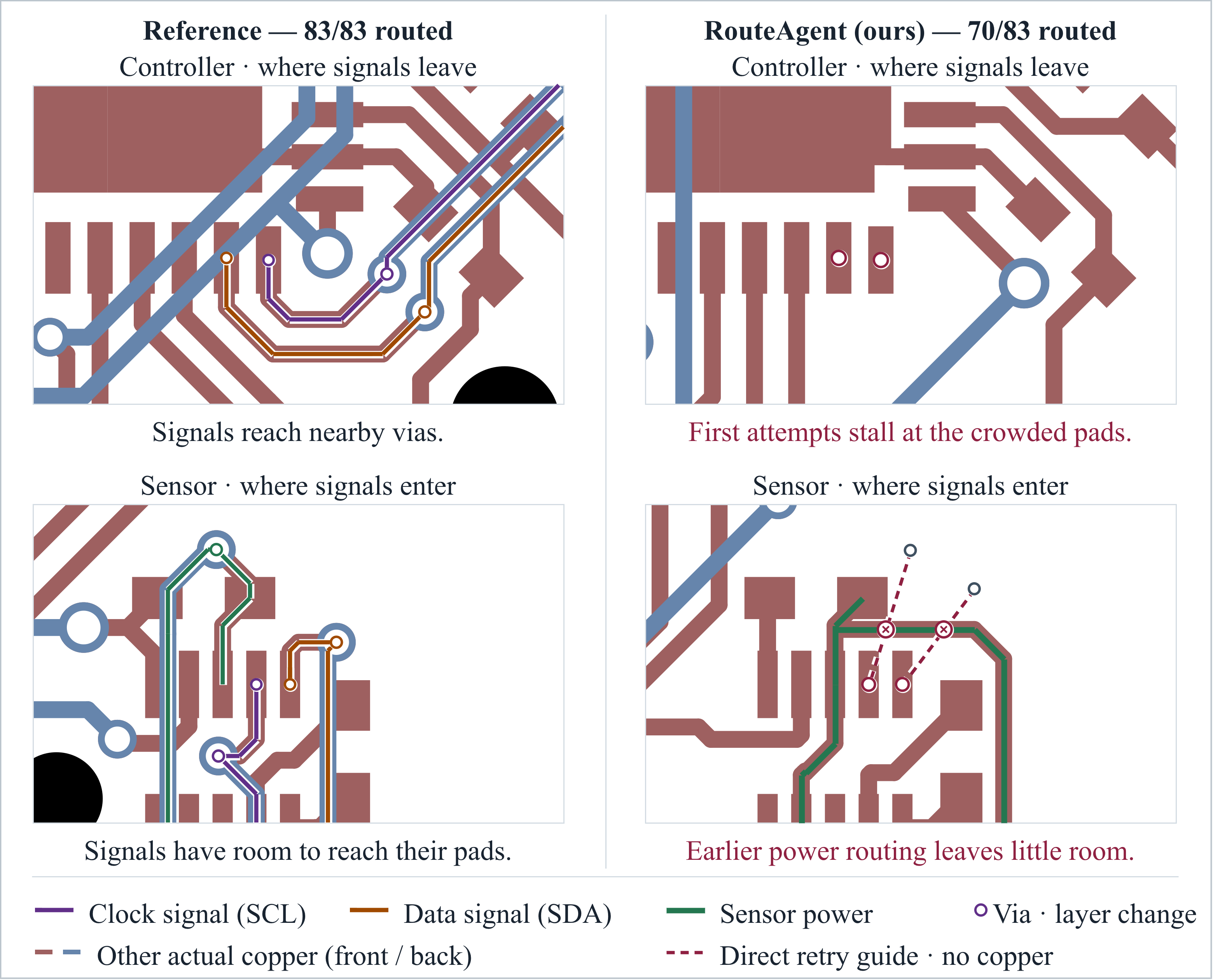}
\caption{Controller and sensor regions in the reference (left) and
PCBRouteAgent output (right). Dashed lines indicate retry guides,
not laid copper.}
\label{fig:routeagent-case-incomplete}
\end{figure}

Earlier routing decisions can consume the local space needed by
later connections. On the D3-C board in
Figure~\ref{fig:routeagent-case-incomplete}, the Opus variant completes
70 of 83 connections, whereas the reference completes all 83.
Two signal nets must connect the controller and the sensor:
SCL carries the clock signal and SDA carries data.

\FloatBarrier

The saved sequence routes sensor power before these signals,
without reserving pad access or space for layer-changing vias.
Initial attempts stall near crowded controller pads. Changing
layers on retry gets the signals out, but their approaches to the
sensor encounter the existing power route. Repair removes
unfinished attempts while retaining that route, so the local
restriction remains. The reference completes both signals with
different paths and layer choices. This case suggests broader
look-ahead in ordering and repair: reserve pad access, prioritize
connections with few alternatives, and revisit earlier routes
when needed.

\FloatBarrier

\subsection{Electrical constraints beyond routing completion}
Full connectivity does not by itself establish electrical usability.
On the D3-B board in Figure~\ref{fig:routeagent-case-electrical},
the Opus variant completes all 55 connections but routes USB
D+ and D- independently.

\FloatBarrier

The two USB traces carry data between the connector and controller
as a differential pair~\cite{siliconlabs2020usb}, for which joint control of spacing and
length balance is recommended.
The reference traces run side by side and each measures 30.25~mm.
PCBRouteAgent produces separate paths of 34.03~mm and 36.49~mm.
The planning input specifies individual net widths and clearances
but does not encode the pair relationship. This exposes a limitation
in signal-specific planning: the task specification and checker
need to capture constraints that depend on signal roles, in
addition to connectivity and basic design rules. Electrical
validation is still required to determine whether this full-speed
USB link operates reliably.

\FloatBarrier

\end{document}